\documentclass[sigconf,natbib=true,anonymous=false]{acmart}
\AtBeginDocument{%
  \providecommand\BibTeX{{%
    \normalfont B\kern-0.5em{\scshape i\kern-0.25em b}\kern-0.8em\TeX}}}

\setcopyright{acmcopyright}
\copyrightyear{2018}
\acmYear{2018}
\acmDOI{XXXXXXX.XXXXXXX}

\acmConference[Conference acronym 'XX]{Make sure to enter the correct
  conference title from your rights confirmation emai}{June 03--05,
  2018}{Woodstock, NY}
\acmPrice{15.00}
\acmISBN{978-1-4503-XXXX-X/18/06}

\usepackage{graphicx}
\usepackage{booktabs}
\usepackage{pifont}
\usepackage{booktabs}
\usepackage{multirow}
\usepackage{amsfonts}
\usepackage{amsmath}
\usepackage{bbm}
\usepackage{bm}
\usepackage{hyperref}

\begin{document}

\title{Investigating Graph Structure Information for Entity Alignment with Dangling Cases}

\author{Jin Xu}
\affiliation{%
    \institution{Shenzhen International Graduate School, Tsinghua University}
    \city{Shenzhen}
    \country{China}
}
\email{xj21@mails.tsinghua.edu.cn}

\author{Yangning Li}
\authornote{Yangning Li contributed equally to this research.}
\affiliation{%
    \institution{Shenzhen International Graduate School, Tsinghua University}
    \city{Shenzhen}
    \country{China}
}
\email{liyn20@mails.tsinghua.edu.cn}

\author{Xiangjin Xie}
\affiliation{%
    \institution{Shenzhen International Graduate School, Tsinghua University}
    \city{Shenzhen}
    \country{China}
}
\email{xxj20@mails.tsinghua.edu.cn}

\author{Yinghui Li}
\affiliation{%
    \institution{Shenzhen International Graduate School, Tsinghua University}
    \city{Shenzhen}
    \country{China}
}
\email{liyinghu20@mails.tsinghua.edu.cn}

\author{Niu Hu}
\affiliation{%
    \institution{Shenzhen International Graduate School, Tsinghua University}
    \city{Shenzhen}
    \country{China}
}
\email{hn20@mails.tsinghua.edu.cn}

\author{Hai-Tao Zheng}
\authornote{Corresponding author}
\affiliation{%
    \institution{Shenzhen International Graduate School, Tsinghua University}
    \city{Shenzhen}
    \country{China}
}
\email{zheng.haitao@sz.tsinghua.edu.cn}

\author{Yong Jiang}
\affiliation{%
    \institution{DAMO Academy, Alibaba Group}
    \city{Hangzhou}
    \country{China}
}
\email{jiangyong.ml@gmail.com}

\renewcommand{\shortauthors}{Trovato and Tobin, et al.}
\newcommand{\lyn}[2][]{{\color{blue}{#2}}}
\begin{abstract}
Entity alignment (EA) aims to discover the equivalent entities in different  knowledge graphs (KGs), which plays an important role in knowledge engineering. Recently,  EA with dangling entities has been proposed as a more realistic setting, which assumes that not all entities have corresponding equivalent entities. In this paper, we focus on this setting. Some work has explored this problem by leveraging translation API, pre-trained word embeddings, and other off-the-shelf tools. However, these approaches over-rely on the side information (e.g., entity names), and fail to work when the side information is absent. On the contrary, they still insufficiently exploit the most fundamental graph structure information in KG. To improve the exploitation of the structural information, we propose a novel entity alignment framework called $\textbf{W}$eakly-$\textbf{O}$ptimal $\textbf{G}$raph $\textbf{C}$ontrastive $\textbf{L}$earning (\textbf{WOGCL}),  which is refined on three dimensions : \textit{(i)} \textbf{Model}. We propose a novel Gated Graph Attention Network to capture local and global graph structure similarity. \textit{(ii)} \textbf{Training}. Two learning objectives: contrastive learning and optimal transport learning are designed to obtain distinguishable entity representations via the optimal transport plan. \textit{(iii)} \textbf{Inference}. In the inference phase, a PageRank-based method is proposed to calculate higher-order structural similarity.
Extensive experiments on two dangling benchmarks demonstrate that our WOGCL outperforms the current state-of-the-art methods with pure structural information in both tranditional (relaxed) and dangling (consolidated) setting. The code will be public soon.
\end{abstract}




\maketitle

\section{Introduction}
\label{section: introduction}
Entity alignment (EA) aims to discover and align equivalent entities in different knowledge graphs (KGs). In real-world scenarios, entity alignment enables the integration of large knowledge graphs which enrich the knowledge graph to facilitate downstream tasks such as question answering, recommendation systems, and search engines \cite{gao2018building}. In recent years, embedding-based entity alignment has seen many attempts \cite{cao2019multi,chen2018co,he2019unsupervised,sun2018bootstrapping,chen2016multilingual} and become the mainstream method. This approach aims to embed the knowledge graph into a low-dimensional space and capture the semantic relevance of entity representations in the embedding space. Then alignments can be discovered through the nearest neighbor search using a distance metric like the Euclidean distance and the cosine distance.

Research on entity alignment has flourished, but an unresolved and significant challenge that existing methods face comes into sight, which is called \emph{dangling entity} problem. \citet{sun2021knowing} pioneer the dangling entity problem. Dangling entities are those entities that exist alone in a KG and cannot find counterparts in another KG. These unmatchable entities would increase the difficulty of EA. Therefore, a more practical problem setting is formally defined where the model needs to detect dangling entities (Dangling Entity Detection, DED) and then align the remaining matchable entities to their counterparts (Entity Alignment, EA). Meanwhile, they emphasize the importance of using structural information for dangling setting entity alignment, rather than relying too much on additional information such as entity names which may introduce name bias \cite{zhang2020industry,liu2020exploring,chaurasiya2022entity,gao2022clusterea} in entity alignment.

 In order to tackle the dangling entity problem and to avoid using biased datasets, a new benchmark dataset $\mathbf{DBP2.0}$ is constructed \cite{sun2021knowing}. DBP2.0 has two remarkable characteristics: (1) A large number of dangling entities are included. In DBP2.0, matchable and dangling entities have similar degree distributions. (2) Only structural information is included in the dataset and other side information (e.g., entity names) is discarded or omitted. The classical translational method MTransE \cite{chen2016multilingual} and the advanced graph neural network method AliNet \cite{sun2020knowledge} are evaluated and proved to perform poorly on DBP2.0.

Recently, several works also attempt to do EA under the dangling setting (Table \ref{tab: categorization}). UED\cite{luo2022accurate} and SoTead\cite{luo2022semi} apply Optimal Transport in the inference stage to tackle the dangling entity problem. However, the performances of these methods are overvalued because of the usage of auxiliary information (entity name) and some off-the-shelf tools (e.g., Glove and Google Translate). On the contrary, structural information is not sufficiently exploited, which is contrary to the motivation of Sun et al. This drawback results in their inability to handle the case where there is no entity name information (e.g., DBP2.0), which greatly reduces the generalizability and effectiveness of the methods.

\begin{table}[htb]
\resizebox{\linewidth}{!}{
\begin{tabular}{@{}lcc@{}}
\toprule
Method  & Additional Tools & Labeled Dangling Entities \\ \midrule
MTransE & \ding{56}        & \ding{52}                       \\
AliNet  & \ding{56}        & \ding{52}                       \\
UED     & Glove, Google Translate      & \ding{56}                       \\
SoTead  & Glove, Google Translate      & \ding{56}                       \\
WOGCL (Ours)    & \ding{56}               & \ding{56}                       \\ \bottomrule
\end{tabular}
}
\caption{Categorization of some EA approaches for dangling cases. MTransE and AliNet utilize labeled dangling entities, while UED and SoTead employ auxiliary information (entity name) and some off-the-shelf tools. WOGCL only uses the most fundamental graph structure information in KG without using additional tools and labeled dangling entities.}
\label{tab: categorization}
\end{table}

In light of these fundamental challenges, to reduce the reliance on non-generic supervisory information and tackle the dangling entity problem, we fully explore the graph structure information in the training and inference stage, which is the basic supervised information for EA models. In this paper, we propose a novel dangling setting entity alignment framework called $\mathbf{W}$eakly-$\mathbf{O}$ptimal $\mathbf{G}$raph $\mathbf{C}$ontrastive $\mathbf{L}$earning $(\mathbf{WOGCL})$. The main contributions are summarized as follows:
\begin{itemize}
    \item $\mathbf{Model.}$ We propose a novel \emph{Gated Graph Attention Network} which models both intra-graph and inter-graph attention mechanisms to intelligently capture local and global relations.
    \item $\mathbf{Training.}$ We propose objective functions based on \emph{Contrastive Learning with Hard Negative Samples} and \emph{Optimal Transport Learning} respectively, where the \emph{Contrastive Learning with Hard Negative Samples} facilitates a more discriminatory representation of entities by hard negative mining and   the \emph{Optimal Transport Learning} (OTL) promotes the entities in two KGs to fit in an optimal allocation.
    \item $\mathbf{Inference.}$ We propose a method for calculating higher-order structural similarity across graphs called \emph{Higher-Order Similarity} (HOS) to help distinguish the alignment of entity pairs more accurately.
\end{itemize}

Empirically, WOGCL outperforms state-of-the-art methods on two dangling benchmarks with pure graph structure information. Furthermore, supplemental experiments demonstrate when combined with entity name information, WOGCL achieves even better performance.

\section{Related Work}
\subsection{Embedding-based Entity Alignment}
Embedding-based entity alignment methods have evolved rapidly and are gradually becoming the mainstream approach of EA in recent years due to their flexibility and effectiveness~\cite{DBLP:journals/corr/abs-2302-08774}, which aim to encode knowledge into low-dimensional embedding space and capture the semantic relatedness of entities\cite{sun2020benchmarking, DBLP:journals/patterns/LiuLTLZ22, DBLP:journals/corr/abs-2211-10997, DBLP:journals/corr/abs-2211-04215}. Numerous current approaches employ translational models such as TransE\cite{bordes2013translating}, TransH\cite{wang2014knowledge}, and TransD\cite{ji2015knowledge} to learn factual knowledge embeddings from relation triples to discover alignment. With the rapid blossoming of graph neural networks, some recent approaches employ GCN\cite{kipf2016semi} and GAT\cite{velickovic2017graph} to capture graph structure information. The GCN-based method GCN-Align \cite{wang2018cross} utilizes the global KG structure to obtain the representation of each entity by aggregating the features of its neighbors. AliNet \cite{sun2020knowledge} employs an attention mechanism and gating mechanism to aggregate both direct and distant neighborhood information. RREA \cite{mao2020mraea} employs relational reflection transformation to obtain relation-specific entity representations~\cite{DBLP:journals/corr/abs-2207-08087}.

Some EA methods\cite{chen2018co,mao2020mraea,sun2018bootstrapping} try to utilize semi-supervised learning to resolve data insufficiency which occurs commonly in EA. Besides, some approaches\cite{cao2019multi,xu2019cross,wu2019jointly,wu2019relation,liu2022selfkg} consider incorporating non-generic side information of entities such as entity attributes and entity names. These auxiliary supervision signals require more additional preprocess or pre-trained multi-language models which may hinder their application in real-world situations and introduce bias to the Web-scale KGs. 

\subsection{Dangling Entity Problem}

Previous practices virtually all assume a one-to-one mapping exists between two KGs, without considering \emph{dangling entity} problem. Sun et al.\cite{sun2021knowing} argues for a dangling entity alignment task and constructs the dataset $\mathbf{DBP2.0}$ from DBpedia. UED\cite{luo2022accurate} mines the literal semantic information to generate pseudo entity pairs and globally guided alignment information for EA and then utilizes the EA results to assist the dangling entity detection. SoTEAD\cite{luo2022semi} sets pseudo entity pairs between KGs based on pre-trained word embeddings~\cite{DBLP:journals/csur/DongLGCLSY23} and conducts contrastive metric learning~\cite{li-etal-2022-past, DBLP:conf/sigir/LiLHYS022} to obtain the transport cost between each entity pair. 

However, these attempts rely heavily on supervised  dangling entity data or pre-trained language models, resulting in dangling entity alignment becoming an additional information-driven task. In this work, we endeavor to investigate the potential of a completely relation-supervised approach without using entity names and dangling entity labels to reduce the cost of entity alignment while improving performance.


\section{Task Definition}
Formally, a KG is defined as $ \mathcal{G}=\left( \mathcal{E},\mathcal{R},\mathcal{T} \right)$, where $\mathcal{E}, \mathcal{R}, \mathcal{T}$ denote the sets of entities, relations and triples respectively. KG stores a large amount of structured information about the real world in the form of triples $ \left\langle head \, entity, relation, tail \, entity  \right\rangle$. In addition, we define $\mathcal{E} = \mathcal{D} \bigcup \mathcal{M}$ where $\mathcal{D}$ denotes a dangling set that contains a large number of entities that have no counterparts and $\mathcal{M}$ denotes matchable set. Given two KGs $\mathcal{G}_1 = \left( \mathcal{E}_1,\mathcal{R}_1,\mathcal{T}_1 \right)$, $\mathcal{G}_2 = \left( \mathcal{E}_2,\mathcal{R}_2,\mathcal{T}_2 \right)$ and a pre-aligned entity pair $ \mathcal{S} = \left\{( e_i, e_j ) \in \mathcal{M} \Vert e_i \equiv e_j \right\} $ where $ \equiv $ indicates equivalence, the EA task seeks to discover the remaining entity alignment.

\section{Methodology}
\begin{figure*}[h]
  \centering
  \includegraphics[width=\linewidth]{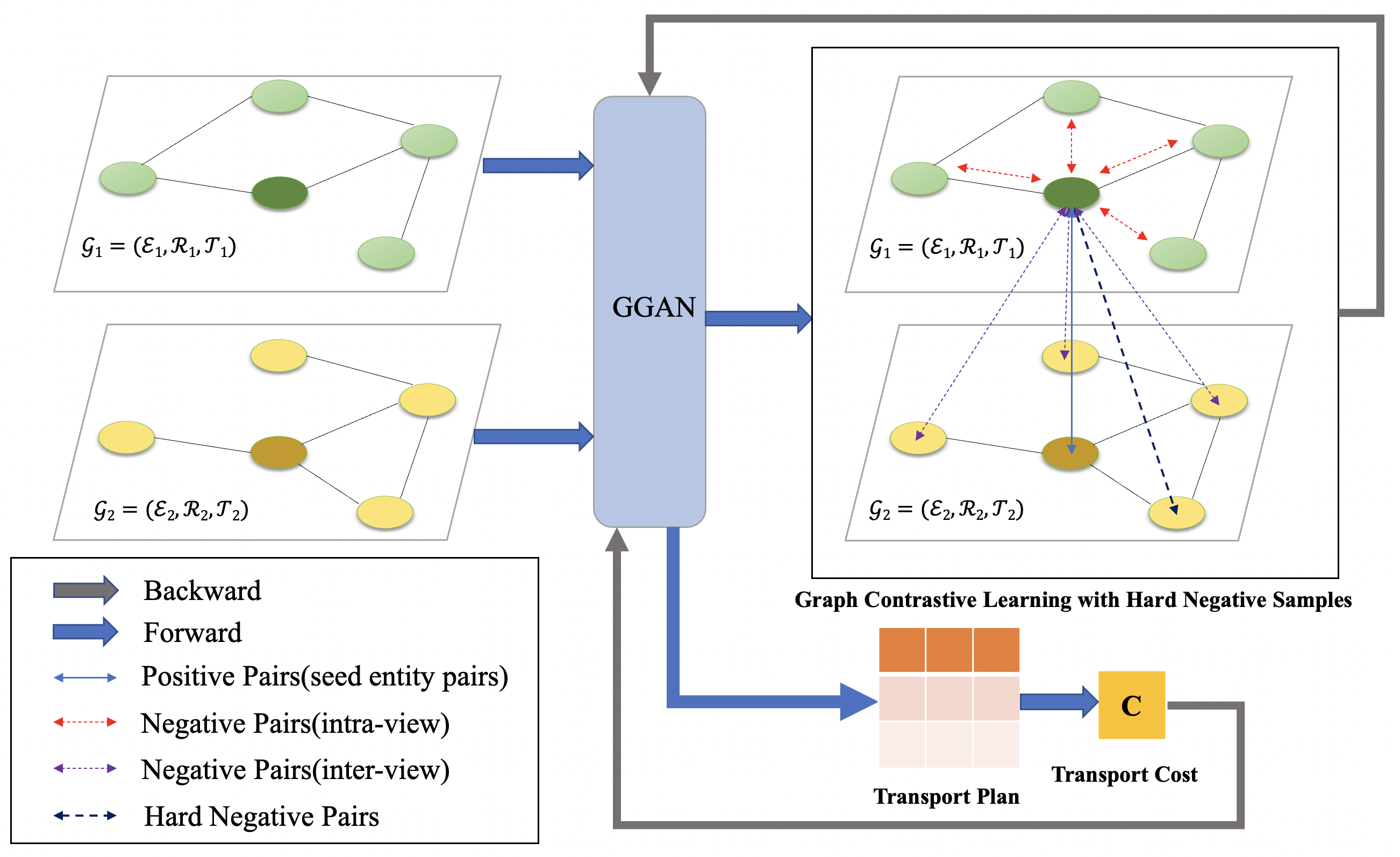}
  \caption{The architecture illustration of Weakly-Optimal Graph Contrastive Learning (WOGCL), composed of graph encoder Gated Graph Attention Network, Contrastive Learning with Hard Negative Samples and Optimal Transport Learning.}
  \label{framework}
  \Description{proposed framework}
\end{figure*}

As mentioned in section \ref{section: introduction}, existing EA methods capture structure information insufficiently and perform less effectively on the dangling setting. To address these defects, we propose \emph{Weakly-Optimal Graph Contrastive Learning} (WOGCL). Figure \ref{framework} depicts that WOGCL is composed of three major components in the model and training: \emph{Gated Graph Attention Network}, \emph{Contrastive Learning with Hard Negative Samples}, and \emph{Optimal Transport Learning}.  \emph{Gated Graph Attention Network} captures local and global relations intelligently via intra-graph and inter-graph attention mechanisms. In the training stage, we leverage \emph{Contrastive Learning with Hard Negative Samples} to refine graph representation. By introducing \emph{Optimal Transport Learning}, we optimize the discrepancy between two distributions of different embeddings. Besides, in the inference stage, which is not shown in Figure \ref{framework}, we propose a novel entity similarity scaling measure \emph{Higher-order Similarity} (HOS) that combines cosine similarity and higher-order structural similarity to solve the dangling entity problem and improve the EA performance. Iterative learning is also adopted for data augmentation. The experimental results show that the proposed method achieves the SOTA in both evaluation settings. In this section, we describe the architecture of WOGCL in detail.

\subsection{Gated Graph Attention Network}
To explicitly model extensive factual information in two KGs, we follow \cite{wang2018cross, sun2020knowledge} and adopt the intra-graph and inter-graph attention mechanism to capture both intra-graph and inter-graph relations smartly in anisotropy graphs. We refer to the model as Gated Graph Attention Network(GGAN) because we employ a gate mechanism to combine intra- and inter-graph representations jointly. 

GNN can be deciphered as a performing message passing mechanism to update node representations while our GGAN can propagate the local and global attention message and update node embeddings. We define two coefficient matrices for both graph views: $\mathbf{C}_{1}$ based on intra-graph attention and $\mathbf{C}_{2}$ based on inter-graph attention. As for the intra-graph attention mechanism, GGAN obtains the node features $ \vec{h}_{e_{i}}$ for entity $e_i \in \mathcal{E} (\mathcal{E}=\{ e_1, e_2, \dots, e_n \})$:
\begin{equation}
\vec{h}_{{e_{i}}}=tanh\left(\sum_{e_{j} \in \mathcal{N}_{e_{i}}} \alpha_{i j} \mathbf{W}_{1} \vec{h}_{e_{j}}\right),
\end{equation}
where $\alpha_{i j}$ represents the intra-graph weight coefficient between entity $e_i$ and $e_j$, $ \mathbf{W}_1 $ represents the trainable transformation matrix, and $N_i$ represents the one-hop neighbors of entity $e_i$ in the graph. For node $e_{j} \in N_{e_{i}}$, the normalized coefficients $\alpha_{i j}$ in $\mathbf{C}_{1}$ computed by the attention mechanism may then be expressed as:
\begin{equation}
\alpha_{i j}=\frac{\exp \left( e_{i j} \right)}{\sum_{e_{k} \in \mathcal{N}_i} \exp \left( e_{ik} \right)},
\end{equation}
where 
\begin{equation}
e_{i j} = \sigma \left( \mathbf{a} \left( \left[\mathbf{W}_{1} \vec{h}_{e_{i}} \| \mathbf{W}_{1} \vec{h}_{e_{j}}\right] \right) \right),
\end{equation}
where $\sigma$ denotes the non-linear activation LeakyReLU, $\mathbf{a} \left( \cdot \right) $ is a shared attentional mechanism:$ \mathbb{R}^F \times \mathbb{R}^F \to \mathbb{R}$ ($F$ is the number of features in each node) and $ \Vert $ is the concatenation operation.

Inter-graph attention aims to model the common subgraph with a similar neighbor community of two KGs to be consistent. In this way, our model can better distinguish which entity pair is equivalent. It is remarkably effective in capturing inter-graph alignment information without introducing additional computational complexity. We define the inter-graph attention coefficient $\theta_{i j}$ in $\mathbf{C}_{2}$ as follows:
\begin{equation}
\theta_{i j}=\frac{\exp \left(\operatorname{sim}\left( \vec{{h}_{e_{i}}}, f \left(\vec{h}_{e_{j}}\right)\right)\right)}{\sum_{e_{k} \in \mathcal{E}_{2}} \exp \left(\operatorname{sim}\left(\vec{h}_{e_{i}}, f \left( \vec{h}_{e_{k}}\right)\right)\right)} ,\\
\end{equation}
where $f (\cdot)$ is the proxy transformation, $\operatorname{sim} \left( \cdot \right)$ is a similarity metric between vectors and in our experiments, the similarity is calculated by inner-product. Thus, we can obtain inter-graph matching features for entity $e_{i}$ by applying the inter-graph attention coefficient $\theta_{i j}$:
\begin{equation}
\vec{h}_{e_{i}}^{\prime}=\text{sigmoid}\left(\mathbf{W}_{2} \vec{h}_{e_{i}}^{\delta}  + \mathbf{b_{2}} \right),
\end{equation}
where
\begin{equation}
\vec{h}_{e_{i}}^{\delta} = \left(\sum_{e_{j} \in \mathcal{E}_{2}} \theta_{i j} \left(\vec{h}_{e_{i}} - f \left(\vec{h}_{e_{j}}\right) \right)\right),
\end{equation}
$\vec{h}_{e_{i}}^{\delta}$ represents the difference between entity $e_{i}$ and all entities in $\mathcal{E}_{2}$ which can more accurately describe the correspondence as well as the semantic difference between entities in heterogeneous KGs.

Intuitively, the resourceful multi-hop neighborhood information can model a single KG locally, whereas the inter-graph attention mechanism can benefit the entity alignment by introducing much more practical global information between two KGs. Therefore, we adopt the gate mechanism\cite{srivastava2015highway} to incorporate these two kinds of graph information to obtain the entity features from diverse views. According to the previous discussion, the combination of intra- and inter-graph information can be formulated as follows:
\begin{equation} \label{gate}
\vec{h}_{e_{i}}^{\text{gate}}=\beta_{e_{i}} \cdot \vec{h}_{e_{i}}+\left(1-\beta_{e_{i}}\right) \cdot \vec{h}_{e_{i}}^{\delta},
\end{equation}
where
\begin{equation}
{\beta}_{e_{i}}=\operatorname{sigmoid}\left(\mathbf{W}_{g} \vec{h}_{e_{i}}^{\prime}+\mathbf{b}_{g}\right),
\end{equation}
where $\mathbf{W}_{g}$ and $\mathbf{b}_{g}$ are trainable gate parameters to be learned. In Equation \eqref{gate}, the first half selects which newly generated information to remember, while the latter selects which information from the past to forget.

\subsection{Contrastive Learning with Hard Negative Samples}
The goal of entity alignment is to be able to find the remaining potentially aligned entity pairs by allowing the model to learn a small portion of the plausible seed pairs. In the case where only structured information$\left( \mathcal{T}, \mathcal{S} \right)$ is available, contrastive learning, especially hard negative sample mining, becomes even more noteworthy. In graph representation learning, translational loss and marginal loss have been widely employed as the embedding objectives. Translational loss can not capture the global information of two KGs while negative sampling in marginal loss is common to select K-nearest neighbors in vector space which is time-consuming.

Inspired by \cite{robinson2020contrastive}, we divide the seed pairs into two different anchors of the same \textit{\textbf{ontology}} and finally derive an effective and efficient hard negative sample mining objective function in entity alignment without using any attribute information to avoid name bias. Given a seed pair $ \left( e_i, e_j \right) \in \mathcal{S}$, these two cross-lingual entities are indeed representations of the same \textit{\textbf{ontology}} in different languages, although they may have different graph structures and dissimilar attributes.

As the relationship between entities and ontologies has been described above,  we assume an underlying set of discrete latent ontologies $ \mathcal{O} $ that represent semantic content and latent equivalence. Let $ m $: $\mathcal{X} \to \mathcal{O} $ be the true underlying hypothesis that assigns latent ontology to inputs. Here we define $ x \Leftrightarrow x^{\prime} $ to denote the ontology equivalence assignments $ m\left( x \right)=m\left(x^{\prime}\right) $. We denote the positive and negative pairs as follows:
\begin{equation}
\mathbb{P}_{pos}=\left\{\left( \mathbf{x}_1,\mathbf{x}_2 \right) \mid m\left( \mathbf{x}_1\right)=m\left(\mathbf{x}_2\right) \right\},
\end{equation}
\begin{equation}
\mathbb{P}_{neg}=\left\{\left( \mathbf{x}_1,\mathbf{x}_2 \right) \mid m\left( \mathbf{x}_1\right) \neq m\left(\mathbf{x}_2\right) \right\}.
\end{equation}
The contrastive loss is then defined as follows:
\begin{equation}
L_{1}=-\sum_{i=1}^{2 N} \log \frac{S_i^{+}}{S_i^{+}+S_i^{-}},
\end{equation}
\begin{equation}
S_i^{+}=e^{\mathbf{z}_i^{\top} \cdot \mathbf{z}_{j(i)} / t},
\end{equation}
\begin{equation}
S_i^{-}=\max \left(\frac{-(2 N-2) \cdot \tau^{+} \cdot S_i^{+}+\widetilde{S_i^{-}}}{1-\tau^{+}}, e^{\frac{-1}{t}}\right),
\end{equation}
\begin{equation}
\widetilde{S_i^{-}}=\frac{(2 N-2) \sum_{k: k \neq i \neq j(i)} e^{(1+\beta) \mathbf{z}_i^{\top} \cdot \mathbf{z}_k / t}}{\sum_{k: k \neq i \neq j(i)} e^{\beta \mathbf{z}_i^{\top} \cdot \mathbf{z}_k / t}},
\end{equation}
where $ S_i^{+}/S_i^{-} $ respectively denotes the similarity between two training samples from equivalent/inequivalent sample pair, $\mathbf{z}$ is the embedding lookup table of all entities in two KGs, $ j\left(i\right) $ represents that entity $i$ and $j$ in the training samples can form a pair of entities containing positive and negative samples, which is to say, $\left( \mathbf{x}_{i}, \mathbf{x}_{j} \right) \in \mathbb{P}_{pos} \cup \mathbb{P}_{neg} $, $\tau^{+}$ is the class-prior probability which can be estimated from data or treated as a hyper-parameter while the first option requires the possession of labeled data before contrastive training and we set $\tau^{+}$ as a hyper-parameter in our experiment, $\beta$ is the concentration parameter which controls the hardness of the negative samples, $t$ is the temperature scaling parameter which is set to 0.5 in our experiments, $N$ is the mini-batch size. It is worth mentioning that the contrastive learning objective function is the main objective function of our embedding module, which helps us to mine the differences between positive and negative samples in the data, especially the hard samples, which play a crucial role in distinguishing whether entity pairs can be aligned or not.
\subsection{Optimal Transport Learning}
To address the dangling entity issue which is common and vital in entity alignment tasks, we promote optimal transport learning for our WOGCL. Let $\mathbf{d}_1$ and $\mathbf{d}_2$ be the discrete distribution of entities from two KGs, then we define the batch-level assignment distance by measuring the difference between $\mathbf{d}_1$ and $\mathbf{d}_2$ with k-Wasserstein distance which is one of the optimal transport distance :

\begin{equation}
\mathcal{C}(\mathbf{d}_1, \mathbf{d}_2)=\inf _{\gamma \in \Pi(\mathbf{d}_1, \mathbf{d}_2)} \mathbb{E}_{(\mathbf{x}_1, \mathbf{x}_2) \sim \gamma} \| \mathbf{x}_1 - \mathbf{x}_2 \|_k ^{k},
\end{equation}
where $ \Pi(\mathbf{d}_1, \mathbf{d}_2) $ is the joint distribution of $\mathbf{d}_1$ and $\mathbf{d}_2$, $\| \mathbf{x}_1 - \mathbf{x}_2 \|_k ^{k}$ denotes the  $L_k$ norm and we set $k$ to 2 in our experiments. However, the infimum to calculate $\mathcal{C}(\mathbf{d}_1, \mathbf{d}_2)$ is highly intractable and thus we adopt Kantorovich-Rubinstein duality points to handle the computation of optimal transport cost. The key idea of Kantorovich is to relax the deterministic nature of transportation, namely the fact that a source point $x_i$ can only be assigned to another point or location $y_{\sigma(i)}$ or $T(x_i)$ only. The Kantorovich relaxation can be expressed by using a permutation $\sigma$ or a bijection map $T: \mathcal{E}_1 \to \mathcal{E}_2$, a coupling matrix $\mathbf{P} \in \mathbb{R}_{+} ^{n \times m}$, where $\mathbf{P}_{i,j}$ describes the amount of mass found at $x_{1i}$ toward $x_{2j}$ in the formalism of discrete distributions. Since $\mathbf{d}_1$ and $\mathbf{d}_2$ can be formulated by a sum of Dirac delta functions as $\mathbf{d}_1=\sum_{i=1}^n a_i\delta_{\mathbf{x}_{1i}}$ and $\mathbf{d}_2=\sum_{j=1}^m b_j\delta_{\mathbf{x}_{2j}}$, then the Kantorovich’s optimal transport problem now reads
\begin{equation}
L_2(\mathbf{d}_1, \mathbf{d}_2) \stackrel{\text { def. }}{=} \min _{\mathbf{P} \in \mathbf{U}(\mathbf{d}_1, \mathbf{d}_2)}\langle\mathbf{C}, \mathbf{P}\rangle \stackrel{\text { def. }}{=} \sum_{i, j} \mathbf{C}_{i, j} \mathbf{P}_{i, j},
\end{equation}where 
\begin{equation}
\mathbf{U}(\mathbf{d}_1, \mathbf{d}_2) \stackrel{\text { def. }}{=}\left\{\mathbf{P} \in \mathbb{R}_{+}^{n \times m}: \mathbf{P} \mathbbm{1}_m=\mathbf{a}, \mathbf{P}^{\mathrm{T}} \mathbbm{1}_n=\mathbf{b}\right\}, 
\end{equation}
where weight vectors $\mathbf{a}=\{ a_i\}_{i=1}^{n} \in \Delta_n$ and $\mathbf{b}=\{ b_i\}_{j=1}^{m} \in \Delta_m$ belong to the $n$- and $m$-dimensional simplex, $\mathbbm{1}_n$ indicates  an $n$-dimensional all-one vector and $\mathbf{P} \mathbbm{1}_n$ can be notated by the following matrix-vector notation as:
\begin{equation}
\mathbf{P} \mathbbm{1}_m=\left(\sum_j \mathbf{P}_{i, j}\right)_i \in \mathbb{R}^n.
\end{equation}
The optimal transport learning measures the discrepancy between each pair of mini-batch group samples from the two cross-lingual KGs.

To better facilitate our model's learning for graph structure and entity distribution, we propose a unified objective function as follows:
\begin{equation}
L = L_1 + \lambda \cdot \omega(t)L_2,
\end{equation}where t is the training epoch, and $\omega(t)$ decreases linearly to 0 as $t$ increases. 
\subsection{Higher-order Structural Information Score}
In order to better utilize only the knowledge graph structure information especially higher-order neighboring information without introducing name bias, we propose a Personalized PageRank(PPR)-based scoring function Higher-order Similarity (HOS) for a small number of seeded graph matching as a measure of whether entity pairs of different KGs can be aligned, motivated by \cite{zhang2021efficient}. Entity alignment is often conducted by exploiting both structural information and semantic features/attributes. When node/edge features are not available, the problem becomes much more challenging. This also poses the challenge of determining whether an entity pair can be aligned or not. As motivated in the introduction, we will employ PPR to quantify the connections between the matchable entities and unmatchable entities which contain the dangling entities.

Given a source vertex $e_{1i} \in \mathcal{E}_1$ and a stopping probability $\alpha$, a decaying random walk is a traversal of $G_1$ that starts from $e_i$ and, at each step: (1) with probability $\left( 1-\alpha \right)$, proceeds to a randomly selected neighbor of the current vertex; or (2) with probability $\alpha$, terminates at the current vertex. For any vertex $e_{1j} \in \mathcal{E}_1$, its PPR value $\pi \left( e_{1i}, e_{2j}\right)$ w.r.t. source $e_{1i}$ is the probability that a decaying random walk from $e_{1i}$ terminates at $e_{1j}$. The same calculation method still holds for another KG.

Starting from vertex $e_{1i}$, let $q_{e_{1(i, j)}}^{(t)}$ be the probability that the (non-decaying) random walk reaches vertex $e_{1j}$ after $t$ steps. Since the probability that a decaying random walk doesn’t terminate before stepping $t$ is $(1-\alpha)^{t}$, the probability that the decaying random walk reaches vertex $e_{1j}$ in the $t^{th}$ step is $p_{e_{1(i, j)}}^{(t)}=(1-\alpha)^{t} \cdot q_{e_{1(i, j)}}^{(t)}$.Therefore,
\begin{equation}
\pi \left( e_{1i}, e_{1j}\right)=\alpha \cdot \sum_{t=0}^{\infty} (1-\alpha)^{t} \cdot q_{e_{1(i, j)}}^{(t)}.
\end{equation}
The main ingredient is a new matching score function, which quantifies the credibility to match any entity pair. Given a small set $S^{\prime} \in S$ of pre-aligned seed pairs, we define the $score \, vector$ of entity  $e_{1i} \in \mathcal{E}_1$ as $T(e_{1i})=\left\{ \pi(s_1, e_{1i}), \pi(s_2, e_{1i}), \dots, \pi(s_k, e_{1i})\right\}$ and that of entity $e_{2j} \in \mathcal{E}_2$ as $T(e_{2j})=\left\{ \pi(s_1, e_{2j}), \pi(s_2, e_{2j}), \dots, \pi(s_k, e_{2j})\right\}$, where $\left(s_{1k}, s_{2k} \right) \in S^{\prime} $ for $1 \le k \le \left|S^{\prime} \right|$. Intuitively, the closer the PPR values of $e_{1i}$ and $e_{2j}$ are (with respect to a seed pair $\left(s_{1k}, s_{2k} \right)$), the more likely entity $e_{1i}$ and $e_{2j}$ are to be a valid match. Thus we propose to define the matching score of two cross-lingual entities $e_{1i}$ and $e_{2j}$ with respect to the $k$-th seed pair $\left(s_{1k}, s_{2k} \right) $ by
\begin{equation}
\mu_k \left( e_{1i}, e_{2j} \right) = \frac{min\left( \pi(s_{1k},e_{1i}),  \pi(s_{2k},e_{2j})\right)}{max\left( \pi(s_{1k},e_{1i}),  \pi(s_{2k},e_{2j})\right)}.
\end{equation}
Then the final matching score can be formulated by:
\begin{equation}
Score_{S} \left( e_{1i}, e_{2j} \right) = \sum_{k=1}^{\left|S^{\prime} \right|}\mu_k \left( e_{1i}, e_{2j} \right).
\end{equation}
This scoring function is able to use the structural information provided by all seeds in $S^{\prime}$ as a more robust global structural information score which can better distinguish the similarities of matchable entities and dangling entities.

Therefore, we define the composite similarity measure of $e_{1i}$ and $e_{2j}$ based on  cosine similarity and higher-order structural information score as:
\begin{equation}
\widetilde{Sim} \left( e_{1i}, e_{2j} \right) = \cos(e_{1i}, e_{2j}) + Score_{S} \left( e_{1i}, e_{2j} \right).
\end{equation}To decrease the similarities between hub entities and other entities, we employ cross-domain similarity local scaling (CSLS) \cite{conneau2017word} to enhance the distance metrics. 
\subsection{Iterative Learning}
In order to better utilize the structural information provided by more unaligned entities to solve the dangling entity problem and to make optimal transport learning more useful, we employ an iterative learning (IL) strategy to obtain a portion of the pseudo-aligned seeds to be added to the next round of training. The goal is to optimize the quality of the pseudo-aligned seed pairs by introducing unaligned pairs, improving the mapping between the spaces of the two KGs, and allowing the test set entities to be mapped from the source entity to the target entity rather than to the dangling entities.

\section{Experiment Setups}

We evaluate WOGCL on two dangling entity alignment datasets: DBP2.0 \cite{sun2021knowing} and MedED \cite{luo2022semi}. DBP2.0 is the dangling entity alignment benchmark and MedED is a cross-lingual medical dangling entity alignment dataset. All the implementations are developed using the PyTorch framework and our experiments are conducted on a workstation with a GeForce GTX 3090Ti GPU and 256GB memory.

\subsection{Datasets}
In this paper, we adopt two different dangling setting entity alignment datasets and conduct extensive experiments to fairly and comprehensively verify the effectiveness and robustness of our framework WOGCL. The details of datasets are shown in Table \ref{tab: datasets}. 

\begin{table}[htb]
\resizebox{\linewidth}{!}{
\begin{tabular}{ccccccc}
\hline
\multicolumn{2}{c}{Datasets} & \# Entities & \# Rel. & \# Triples & \# Align.                & \#Dang. \\ \hline
\multirow{2}{*}{DBP2.0} & ZH & 84,996      & 3,706   & 286,067    & \multirow{2}{*}{33,183}  & 51,813  \\
                        & EN & 118,996     & 3,402   & 586,868    &                          & 85,813  \\
\multirow{2}{*}{DBP2.0} & JA & 100,860     & 3,243   & 347,204    & \multirow{2}{*}{39,770}  & 61,090  \\
                        & EN & 139,304     & 3,396   & 668,341    &                          & 99,534  \\
\multirow{2}{*}{DBP2.0} & FR & 221,327     & 2,841   & 802,678    & \multirow{2}{*}{123,952} & 97,375  \\
                        & EN & 278,411     & 4,598   & 1,287,231  &                          & 154,459 \\
\multirow{2}{*}{MedED}  & ES & 19,639      & 537     & 610,488    & \multirow{2}{*}{11,298}   & 8,341   \\
                        & EN & 18,855      & 614     & 855,466    &                          & 7,557   \\
\multirow{2}{*}{MedED}  & FR & 19,109      & 431     & 449,276    & \multirow{2}{*}{8,365}    & 12,674  \\
                        & EN & 18,855      & 614     & 855,466    &                          & 12,420  \\ \hline
\end{tabular}
}
\caption{Statistics of DBP2.0 and MedED.}
\label{tab: datasets}
\end{table}

\subsection*{DBP2.0}
DBP2.0 is an entity alignment dataset with abundant dangling entities constructed using multilingual knowledge from the Infobox Data of DBpedia\cite{lehmann2015dbpedia}. The dataset contains three pairs of cross-lingual KGs, ZH-EN (Chinese to English), JA-EN (Japanese to English), and FR-EN (French to English). Only relation triples are built into the dataset, which is our starting point, i.e., finding entity alignment and detecting dangling entities solely with graph structural similarity. Besides, matchable and dangling entities have similar degree distributions which fit the characteristics of the knowledge graph in practice. We follow the dataset splits in \cite{sun2021knowing} where 30\% dangling entities are for training, 20\% for validation, and 50\% for testing. 

\subsection*{MedED}
MedED is constructed from the Unified Medical Language System (UMLS)\cite{bodenreider2004unified}. The concepts in UMLS have several terms in different languages. MedED has not been released and we follow the methods proposed from \cite{luo2022semi} to extract the KGs from the large-scale resource successfully. We extracted the KGs of three different languages, i.e., English, French, and Spanish, and then constructed the KG pairs of FR-EN (French to English) and ES-EN (Spanish to English). We select 20000 entities in the UMLS with the most relation triples for the specified language, and then delete entities unrelated to other selected entities. To keep the same dataset splits, we split 30\% of entity pairs as the training set, 20\% as the validation set, and 50\% as the test set. 

\subsection{Compared methods}
To the best of our knowledge, the only method proposed for the benchmark  DBP2.0 is \cite{sun2021knowing}. For a fair comparison with it, we choose MTransE \cite{chen2016multilingual} and AliNet \cite{sun2020knowledge} as our baseline models. Moreover, three techniques are proposed to resolve the dangling entity problem, i.e., nearest neighbor(NN) classification, marginal ranking (MR), and background ranking (BR). As MR and BR perform much better than NN classification, we combine the baseline models with MR and BR as our baselines. For MedED, \cite{luo2022semi} and \cite{luo2022accurate} utilize pre-trained language models to generate pseudo labels to achieve unsupervised learning while we only employ relation triples and a small set of pre-aligned entity pairs to do EA and DED. Thus, we choose the MTransE with MR as the baseline for MedED.

\subsection{Evaluation protocals}
In this paper, we use two evaluation settings as recommended by Sun et al.\cite{sun2021knowing}, i.e., \emph{consolidated evaluation} and \emph{relaxed evaluation}. The first one is to conduct dangling entity detection and alignment search for matchable entities involving dangling entities, while the other one is a simplified setting and seeks to find alignments between matchable entities. For the \emph{consolidated evaluation}, we evaluate the performance of entity alignment and dangling entity detection using precision, recall, and F1 score following \cite{sun2021knowing}. With regards to \emph{relaxed evaluation}, we use mean reciprocal rank (MRR), Hits@1, and Hits@10 as metrics as most works do. The Hits@k score is estimated by measuring the proportion of correctly aligned pairs in the top-k ranking list, e.g., the Hits@1 score indicates the portion of the targets in the test set that have been correctly ranked in the top-1. The higher metric values indicate better performance.

\subsection{Implementation Details}
For KG embeddings and model weights in our framework, we employ Xavier initialization \cite{glorot2010understanding} and then optimize them using RMSProp optimizer. For all datasets, we use essentially the same config: The depth of GNN is 2; attention-heads is 1; the class-prior probability $\tau^{+}$ = 0.1; the hardness coefficient $\beta = 1$; the temperature scaling parameter $t = 0.5$; the distance metric in Optimal Transport Learning is set to $L_{2}$ norm; balance coefficient $\lambda$ = 0.3; learning rate (lr) is set to 0.005; batch size is 512; epoch is 20; training turns is 5; dropout rate is set to 30\%. For DBP2.0, the hidden size of node and relation are both set to 32 because of its large number of entities and relations, while the hidden size of node and relation is set to 128 for MedED.

\begin{table*}[hbt!]
\centering
\resizebox{\linewidth}{!}{
\begin{tabular}{lccccccccccccccc}
\hline
\multirow{2}{*}{Methods} & \multicolumn{3}{c}{DBP2.0$_{ZH-EN}$}                  & \multicolumn{3}{c}{DBP2.0$_{JA-EN}$ }                 & \multicolumn{3}{c}{DBP2.0$_{FR-EN}$}                 & \multicolumn{3}{c}{MedED$_{ES-EN}$}                   & \multicolumn{3}{c}{MedED$_{FR-EN}$}                   \\ \cline{2-16} 
                         & H@1            & H@10           & MRR            & H@1            & H@10           & MRR            & H@1            & H@10           & MRR            & H@1            & H@10           & MRR            & H@1            & H@10           & MRR            \\ \hline
Alinet                   & 0.332          & 0.594          & 0.421          & 0.338          & 0.596          & 0.429          & 0.223          & 0.473          & 0.306          & -              & -              & -              & -              & -              & -              \\
Alinet+MR                & 0.343          & 0.606          & 0.433          & 0.349          & 0.608          & 0.438          & 0.230          & 0.477          & 0.312          & -              & -              & -              & -              & -              & -              \\
Alinet+BR                & 0.333          & 0.599          & 0.426          & 0.341          & 0.608          & 0.431          & 0.214          & 0.468          & 0.298          & -              & -              & -              & -              & -              & -              \\ \hline
MTransE                  & 0.358          & 0.675          & 0.463          & 0.348          & 0.661          & 0.453          & 0.245          & 0.524          & 0.338          & -              & -              & -              & -              & -              & -              \\
MTransE+MR               & 0.378          & 0.693          & 0.487          & 0.373          & 0.686          & 0.476          & 0.259          & 0.541          & 0.348          & 0.815          & 0.961          & 0.869          & 0.708          & 0.955          & 0.800          \\
MTransE+BR               & 0.360          & 0.678          & 0.468          & 0.344          & 0.660          & 0.451          & 0.251          & 0.525          & 0.342          & -              & -              & -              & -              & -              & -              \\ \hline
WOGCL (Basic)                   & \textbf{0.578} & \textbf{0.875} & \textbf{0.679} & \textbf{0.556} & \textbf{0.860} & \textbf{0.660} & \textbf{0.367} & \textbf{0.687} & \textbf{0.470} & \textbf{0.939} & \textbf{0.992} & \textbf{0.960} & \textbf{0.940} & \textbf{0.992} & \textbf{0.961} \\
WOGCL (Semi)              & 0.664          & 0.892          & 0.744          & 0.649          & 0.872          & 0.729          & 0.419          & 0.709          & 0.516          & 0.961          & 0.995          & 0.974          & 0.976          & 0.994          & 0.985          \\ \hline
\end{tabular}
}
\caption{Entity alignment results (relaxed evaluation) on DBP2.0 and MedED.}
\label{tab: relaxed setting}
\end{table*}

\begin{table*}[hbt!]
\resizebox{\linewidth}{!}{
\begin{tabular}{lccccccccccccccc}
\hline
\multirow{2}{*}{Methods} & \multicolumn{3}{c}{DBP2.0$_{ZH-EN}$}                  & \multicolumn{3}{c}{DBP2.0$_{JA-EN}$}                  & \multicolumn{3}{c}{DBP2.0$_{FR-EN}$}                  & \multicolumn{3}{c}{MedED$_{ES-EN}$}                   & \multicolumn{3}{c}{MedED$_{FR-EN}$}                   \\ \cline{2-16} 
                         & Prec.          & Rec.           & F1             & Prec.          & Rec.           & F1             & Prec.          & Rec.           & F1             & Prec.          & Rec.           & F1             & Prec.          & Rec.           & F1             \\ \hline
Alinet+MR                & 0.207          & 0.299          & 0.245          & 0.231          & 0.321          & 0.269          & 0.195          & 0.190          & 0.193          & -              & -              & -              & -              & -              & -              \\
Alinet+BR                & 0.203          & 0.286          & 0.238          & 0.223          & 0.306          & 0.258          & 0.183          & 0.181          & 0.182          & -              & -              & -              & -              & -              & -              \\ \hline
MTransE+MR               & 0.302          & 0.349          & 0.324          & 0.313          & 0.367          & 0.338          & 0.260          & 0.220          & 0.238          & 0.770          & 0.660          & 0.711          & 0.628          & 0.677          & 0.652          \\
MTransE+BR               & 0.312          & 0.362          & 0.335          & 0.314          & 0.363          & 0.336          & 0.265          & 0.208          & 0.233          & -              & -              & -              & -              & -              & -              \\ \hline
WOGCL (Basic)                   & \textbf{0.387} & \textbf{0.393} & \textbf{0.390} & \textbf{0.417} & \textbf{0.456} & \textbf{0.435} & \textbf{0.393} & \textbf{0.289} & \textbf{0.333} & \textbf{0.951} & \textbf{0.911} & \textbf{0.931} & \textbf{0.905} & \textbf{0.897} & \textbf{0.901} \\
WOGCL (Semi)              & 0.485          & 0.619          & 0.542          & 0.491          & 0.674          & 0.567          & 0.430          & 0.434          & 0.432          & 0.964          & 0.960          & 0.962          & 0.932          & 0.976          & 0.953          \\ \hline
\end{tabular}
}
\caption{Entity alignment results (consolidated evaluation) on DBP2.0 and MedED. MR refers to marginal ranking and BR refers to background ranking.}
\label{tab: consolidated setting}
\end{table*}

\begin{table*}[hbt!]
\resizebox{\linewidth}{!}{
\begin{tabular}{lccccccccccccccc}
\hline
\multirow{2}{*}{Methods} & \multicolumn{3}{c}{DBP2.0$_{ZH-EN}$}                                               & \multicolumn{3}{c}{DBP2.0$_{JA-EN}$}                                               & \multicolumn{3}{c}{DBP2.0$_{FR-EN}$}                                               & \multicolumn{3}{c}{MedED$_{ES-EN}$}                                                & \multicolumn{3}{c}{MedED$_{FR-EN}$}                                                \\ \cline{2-16} 
                         & \multicolumn{1}{l}{Prec.} & \multicolumn{1}{l}{Rec.} & \multicolumn{1}{l}{F1} & \multicolumn{1}{l}{Prec.} & \multicolumn{1}{l}{Rec.} & \multicolumn{1}{l}{F1} & \multicolumn{1}{l}{Prec.} & \multicolumn{1}{l}{Rec.} & \multicolumn{1}{l}{F1} & \multicolumn{1}{l}{Prec.} & \multicolumn{1}{l}{Rec.} & \multicolumn{1}{l}{F1} & \multicolumn{1}{l}{Prec.} & \multicolumn{1}{l}{Rec.} & \multicolumn{1}{l}{F1} \\ \hline
Alinet+MR                & 0.752                     & 0.538                    & 0.627                  & 0.779                     & 0.580                    & 0.665                  & 0.552                     & 0.570                    & 0.561                  & -                         & -                        & -                      & -                         & -                        & -                      \\
Alinet+BR                & 0.762                     & 0.556                    & 0.643                  & 0.783                     & 0.591                    & 0.673                  & 0.547                     & 0.556                    & 0.552                  & -                         & -                        & -                      & -                         & -                        & -                      \\
MTransE+MR               & 0.781                     & 0.702                    & 0.740                  & 0.799                     & 0.708                    & 0.751                  & 0.482                     & 0.575                    & 0.524                  & 0.684                     & 0.831                    & 0.751                  & 0.864                     & 0.814                    & 0.838                  \\
MTransE+BR               & \textbf{0.811}            & 0.728                    & \textbf{0.767}         & \textbf{0.816}            & 0.733                    & \textbf{0.772}         & 0.539                     & \textbf{0.686}           & 0.604                  & -                         & -                        & -                      & -                         & -                        & -                      \\
WOGCL                    & 0.672                     & \textbf{0.748}           & 0.708                  & 0.751                     & \textbf{0.784}           & 0.767                  & \textbf{0.745}            & 0.588                    & \textbf{0.657}         & \textbf{0.926}            & \textbf{0.861}           & \textbf{0.892}         & \textbf{0.922}            & \textbf{0.907}           & \textbf{0.914}         \\ \hline
\end{tabular}
}
\caption{Dangling entity detection results on DBP2.0 and MedED.}
\label{tab: dangling detection}
\end{table*}

\begin{table*}[hbt!]
\resizebox{\linewidth}{!}{
\begin{tabular}{c|lccccccccccccccc}
\hline
                        & \multicolumn{1}{c|}{}                         & \multicolumn{3}{c}{DBP2.0$_{ZH-EN}$}                  & \multicolumn{3}{c}{DBP2.0$_{JA-EN}$}                  & \multicolumn{3}{c}{DBP2.0$_{FR-EN}$}                  & \multicolumn{3}{c}{MedED$_{ES-EN}$}                   & \multicolumn{3}{c}{MedED$_{FR-EN}$}                                                 \\ \cline{3-17} 
                        & \multicolumn{1}{c|}{\multirow{-2}{*}{Models}} & H@1 & H@10 & MRR   & H@1 & H@10 & MRR   & H@1 & H@10 & MRR   & H@1 & H@10 & MRR   & H@1 & H@10 & MRR   \\ \hline
                        & \multicolumn{1}{l|}{WOGCL}                                         & 0.578                            & 0.875                             & 0.679 & 0.556                            & 0.860                             & 0.660 & 0.367                            & 0.687                             & 0.470 & 0.939                            & 0.992                             & 0.960 & 0.940                            & 0.992                             & 0.961 \\
                        & \multicolumn{1}{l|}{\quad w/o OTL}                                       & 0.553                            & 0.865                             & 0.658 & 0.536                            & 0.851                             & 0.643 & 0.361                            & 0.691                             & 0.467 & 0.921                            & 0.990                             & 0.948 & 0.933                            & 0.990                             & 0.956 \\
\multirow{-3}{*}{Basic} & \multicolumn{1}{l|}{\quad w/o HOS}                                       & 0.571                            & 0.875                             & 0.674 & 0.548                            & 0.855                             & 0.654 & 0.364                            & 0.680                             & 0.466 & 0.930                            & 0.992                             & 0.954 & 0.932                            & 0.992                             & 0.956 \\ \hline
                        & \multicolumn{1}{l|}{WOGCL}                                         & 0.664                            & 0.892                             & 0.744 & 0.649                            & 0.872                             & 0.729 & 0.419                            & 0.709                             & 0.516 & 0.961                            & 0.995                             & 0.974 & 0.976                            & 0.994                             & 0.985 \\
                        & \multicolumn{1}{l|}{\quad w/o OTL}                                       & 0.640                            & 0.882                             & 0.726 & 0.624                            & 0.866                             & 0.711 & 0.416                            & 0.697                             & 0.513 & 0.952                            & 0.993                             & 0.969 & 0.972                            & 0.995                             & 0.982 \\
\multirow{-3}{*}{Semi}  & \multicolumn{1}{l|}{\quad w/o HOS}                                       & 0.658                            & 0.890                             & 0.740 & 0.640                            & 0.872                             & 0.724 & 0.399                            & 0.645                             & 0.474 & 0.955                            & 0.994                             & 0.971 & 0.975                            & 0.995                             & 0.983 \\ \hline
\end{tabular}
}
\caption{Entity alignment ablation results (relaxed evaluation) on DBP2.0.}
\label{tab: DBP2.0 ablation relaxed setting}
\end{table*}

\begin{table*}[hbt!]
\resizebox{\linewidth}{!}{
\begin{tabular}{c|lccccccccccccccc}
\hline
                        & \multicolumn{1}{c|}{}                         & \multicolumn{3}{c}{DBP2.0$_{ZH-EN}$}                  & \multicolumn{3}{c}{DBP2.0$_{JA-EN}$}                  & \multicolumn{3}{c}{DBP2.0$_{FR-EN}$}                  & \multicolumn{3}{c}{MedED$_{ES-EN}$}                   & \multicolumn{3}{c}{MedED$_{FR-EN}$}                                                                                                        \\ \cline{3-17} 
                        & \multicolumn{1}{c|}{\multirow{-2}{*}{Models}} & \multicolumn{1}{l}{{\color[HTML]{000000} Prec.}} & \multicolumn{1}{l}{{\color[HTML]{000000} Rec.}} & \multicolumn{1}{l}{{\color[HTML]{000000} F1}} & \multicolumn{1}{l}{{\color[HTML]{000000} Prec.}} & \multicolumn{1}{l}{{\color[HTML]{000000} Rec.}} & \multicolumn{1}{l}{{\color[HTML]{000000} F1}} & \multicolumn{1}{l}{{\color[HTML]{000000} Prec.}} & \multicolumn{1}{l}{{\color[HTML]{000000} Rec.}} & \multicolumn{1}{l}{{\color[HTML]{000000} F1}} & \multicolumn{1}{l}{{\color[HTML]{000000} Prec.}} & \multicolumn{1}{l}{{\color[HTML]{000000} Rec.}} & \multicolumn{1}{l}{{\color[HTML]{000000} F1}} & \multicolumn{1}{l}{{\color[HTML]{000000} Prec.}} & \multicolumn{1}{l}{{\color[HTML]{000000} Rec.}} & \multicolumn{1}{l}{{\color[HTML]{000000} F1}} \\ \hline
                        & \multicolumn{1}{l|}{WOGCL}                                         & 0.387                                            & 0.393                                           & 0.390                                         & 0.417                                            & 0.456                                           & 0.435                                         & 0.393                                            & 0.289                                           & 0.333                                         & 0.951                                            & 0.911                                           & 0.931                                         & 0.905                                            & 0.897                                           & 0.901                                         \\
                        & \multicolumn{1}{l|}{\quad w/o OTL}                                       & 0.374                                            & 0.377                                           & 0.376                                         & 0.404                                            & 0.436                                           & 0.420                                         & 0.389                                            & 0.281                                           & 0.326                                         & 0.934                                            & 0.880                                           & 0.906                                         & 0.885                                            & 0.877                                           & 0.881                                         \\
\multirow{-3}{*}{Basic} & \multicolumn{1}{l|}{\quad w/o HOS}                                       & 0.379                                            & 0.386                                           & 0.382                                         & 0.410                                            & 0.443                                           & 0.426                                         & 0.389                                            & 0.289                                           & 0.331                                         & 0.942                                            & 0.893                                           & 0.917                                         & 0.900                                            & 0.885                                           & 0.892                                         \\ \hline
                        & \multicolumn{1}{l|}{WOGCL}                                         & 0.485                                            & 0.619                                           & 0.542                                         & 0.491                                            & 0.674                                           & 0.567                                         & 0.430                                            & 0.434                                           & 0.432                                         & 0.964                                            & 0.960                                           & 0.962                                         & 0.932                                            & 0.976                                           & 0.953                                         \\
                        & \multicolumn{1}{l|}{\quad w/o OTL}                                       & 0.450                                            & 0.589                                           & 0.502                                         & 0.464                                            & 0.630                                           & 0.535                                         & 0.421                                            & 0.419                                           & 0.420                                         & 0.951                                            & 0.939                                           & 0.945                                         & 0.913                                            & 0.965                                           & 0.939                                         \\
\multirow{-3}{*}{Semi}  & \multicolumn{1}{l|}{\quad w/o HOS}                                       & 0.475                                            & 0.609                                           & 0.534                                         & 0.491                                            & 0.669                                           & 0.561                                         & 0.414                                            & 0.430                                           & 0.422                                         & 0.953                                            & 0.950                                           & 0.951                                         & 0.932                                            & 0.975                                           & 0.953                                         \\ \hline
\end{tabular}
}
\caption{Entity alignment ablation results (consolidated evaluation) on DBP2.0.}
\label{tab: DBP2.0 ablation consolidated setting}
\end{table*}

\begin{table*}[hbt!]
\resizebox{\linewidth}{!}{
\begin{tabular}{cccccc}
\hline
Source Entity                      & {\color[HTML]{FE0000} NN Entity(dangling entity)}                                 & Cosine Sim. & {\color[HTML]{000000} \textbf{Target Entity}}                          & Cosine Sim. & Cosine Sim.(W OTL) \\ \hline
eosinophilia                       & {\color[HTML]{FE0000} stries ovariennes}                                          & 0.8655      & {\color[HTML]{000000} \textbf{éosinophilie}}                           & 0.8538      & \textbf{0.9272}    \\
shoulder dislocation               & {\color[HTML]{FE0000} luxation ouverte du coude, site non précisé}                & 0.8026      & {\color[HTML]{000000} \textbf{luxation de l'épaule}}                   & 0.7855      & \textbf{0.8688}    \\
adenocarcinoma of appendix         & {\color[HTML]{FE0000} carcinome de l'appendice}                                   & 0.8695      & {\color[HTML]{000000} \textbf{adénocarcinome de l'appendice}}          & 0.866       & \textbf{0.8929}    \\
crohn's disease of small intestine & {\color[HTML]{FE0000} entérite régionale de l'intestin grêle et du gros intestin} & 0.9164      & {\color[HTML]{000000} \textbf{entérite régionale de l'intestin grêle}} & 0.8847      & \textbf{0.9405}    \\
hallucinations                     & {\color[HTML]{FE0000} agitation}                                                  & 0.8838      & {\color[HTML]{000000} \textbf{hallucinations}}                         & 0.8689      & \textbf{0.9283}    \\
basophils                          & {\color[HTML]{FE0000} promyélocytes}                                              & 0.8992      & {\color[HTML]{000000} \textbf{granulocytes basophiles}}                & 0.8957      & \textbf{0.9248}    \\
pustule                            & {\color[HTML]{FE0000} rash pustuleux}                                             & 0.8671      & {\color[HTML]{000000} \textbf{pustule}}                                & 0.8554      & \textbf{0.8907}    \\ \hline
\end{tabular}
}
\caption{Some dangling target entities are wrongly predicted as matchable, while Optimal Transport Learning (OTL) helps WOGCL predict the right target entity with higher probabilities. NN Entity(dangling entity) denotes the dangling target entities that are wrongly predicted as matchable entities by nearest neighbor searching. The last column denotes the cosine similarity between some matchable pairs becomes highest with OTL.}
\label{tab: case study}
\end{table*}

\section{Experiment Results}

In this section, we report our experiments to show the effectiveness of WOGCL. We present the results in two alignment settings separately in Sec. \ref{subsection:Relaxed evaluation} and \ref{subsection:Consolidated evaluation}. We conduct an ablation study and demonstrate that each component of WOGCL has a role to play in Sec. \ref{subsection:Ablation Study}, followed by a case study to show the importance of optimal transport learning in Sec. \ref{subsection:Case study}.

\subsection{Relaxed evaluation}
\label{subsection:Relaxed evaluation}
In table \ref{tab: relaxed setting}, we report the performances of relaxed evaluation on DBP2.0 and MedED. Our method consistently achieves the best performance across all datasets. On the large-scale dataset DBP2.0, WOGCL outperforms other methods by at least 20\% in terms of both Hits@1 and MRR. On MedED, the performances are increased by at least 11\% on Hits@1 and MRR compared to MTransE with MR. It is straightforward to find from the experimental results that our WOGCL is capable of capturing the structural information of the graph structure nicely and obtaining sufficiently discriminative entity representation. Benefiting from the iterative learning strategy to generate more labeled data for the next training turn, the overall performances of the semi-supervised methods exceed the basic methods greatly. We believe the iterative learning strategy is uncomplicated and effective. All these experimental results demonstrate the effectiveness of WOGCL in capturing structural information.

\subsection{Consolidated evaluation}
\label{subsection:Consolidated evaluation}

\noindent \textbf{Dangling entity detection.} According to the results in Table \ref{tab: dangling detection}, WOGCL essentially achieves results close to the corresponding baseline by Sun et al. \cite{sun2021knowing} on DBP2.0 while WOGCL consistently achieves much better F1 scores compared with the corresponding baseline on MedED. With better precision and F1 scores on FR-EN of DBP2.0, WOGCL has the same level or slightly worse precision and F1 scores on ZH-EN and JA-EN compared with baselines. For the DED task on MedED, the proposed method focuses more on the precision in recognizing dangling entities. In summary, WOGCL demonstrates superior effectiveness for detecting dangling entities.

\noindent \textbf{Entity alignment with dangling entities.} Table \ref{tab: consolidated setting} shows the results of consolidated evaluation on DBP2.0 and MedED. In general, basic WOGCL consistently presents better F1 scores than baseline methods. The relative improvement of DBP2.0 ranges from 16\% to 43\%, while the performances are increased ranges from 7\% to 24\% on MedED. Compared to the basic version, the semi-supervised strategy greatly improves the performances of EA on DBP2.0 and MedED. These results indicate that WOGCL successfully uses pre-aligned entity pairs to lead to the more accurate mapping of entities across the knowledge graph to alleviate the misleading influence of dangling entities. Moreover, these results demonstrate that WOGCL successfully uses DED with high precision to reduce the scope of EA and enhance the performance of EA.

\subsection{Ablation study}
\label{subsection:Ablation Study}
To investigate the effectiveness of each module in WOGCL, we conduct an ablation study on both evaluation settings and show relaxed evaluation and consolidated evaluation results in Tab. \ref{tab: DBP2.0 ablation relaxed setting} and Tab. \ref{tab: DBP2.0 ablation consolidated setting}. We present the ablation study for WOGCL on both DBP2.0 and MedED, including the ablation of Optimal Transport Learning (OTL) and Higher-order Similarity (HOS). Besides, Table \ref{tab: relaxed literal ablation} and \ref{tab: consolidated literal ablation} provide WOGCL variants with literal information in the relaxed and consolidated settings. Our main observations are: (1) The OTL is stable and effective, causing significant improvements on Hits@1, MRR for relaxed evaluation, and F1 scores for consolidated evaluation compared with WOGCL without OTL. In semi-supervised methods, optimal transport learning prevents performance degradation due to the introduction of wrongly labeled entity pairs. (2)
Ablating the HOS leads to a slight decrease in most cases, indicating they are helpful in the inference stage. These ablation experiments show that the method is significant and has improved remarkably. 

\begin{table}[hbt!]
\begin{tabular}{lcccccc}
\hline
          \multirow{2}{*}{Methods} & \multicolumn{3}{c}{MedED$_{ES-EN}$} & \multicolumn{3}{c}{MedED$_{FR-EN}$} \\
         & H@1      & H@10     & MRR      & H@1      & H@10     & MRR      \\ \hline
Init-Emb & 0.740    & 0.889    & 0.791    & 0.752    & 0.881    & 0.798    \\ \hline
WOGCL    & 0.939    & 0.992    & 0.960    & 0.940    & 0.992    & 0.961    \\
\quad w EN  & \textbf{0.959}    & \textbf{0.997}    & \textbf{0.974}    & \textbf{0.958}    & \textbf{0.998}    & \textbf{0.974}    \\ \hline
\end{tabular}
\caption{Ablation Study of literal information on MedED (relaxed evaluation). EN refers to entity name.}
\label{tab: relaxed literal ablation}
\end{table}

\begin{table}[hbt!]
\begin{tabular}{lcccccc}
\hline
\multirow{2}{*}{Methods} & \multicolumn{3}{c}{MedED$_{ES-EN}$}                   & \multicolumn{3}{c}{MedED$_{FR-EN}$}                   \\
                         & Prec.          & Rec.           & F1             & Prec.          & Rec.           & F1             \\ \hline
Init-Emb                 & 0.904          & 0.680          & 0.776          & 0.846          & 0.667          & 0.746          \\ \hline
WOGCL                    & 0.951          & 0.911          & 0.931          & \textbf{0.905}          & 0.897          & 0.901          \\
\quad w EN                  & \textbf{0.955} & \textbf{0.933} & \textbf{0.944} & 0.902 & \textbf{0.922} & \textbf{0.912} \\ \hline
\end{tabular}
\caption{Ablation Study of literal information on MedED (consolidated evaluation).}
\label{tab: consolidated literal ablation}
\end{table}

Although our motivation is to use graph structure information for entity alignment in different settings, it can be found in Table \ref{tab: relaxed literal ablation} and \ref{tab: consolidated literal ablation} that our WOGCL has a strong generalization ability and portability, which can be perfectly combined with literal information (entity name) and get performance improvements. We adopt a state-of-the-art multi-lingual pre-trained language model LaBSE \cite{feng2020language} to get the initial embedding of entity. Despite the maturity of pre-trained models, over-reliance on the use of name information when performing entity alignment can instead lead to performance degradation due to the bias introduced by name information. By comparing the basic WOGCL with the Init-Emb approach, it is easy to see the superiority of our model in discovering entity alignment using structural similarity. 

\subsection{Case study}
\label{subsection:Case study}
To further investigate the superiority of OTL, we provide a case study on FR-EN of MedED comparing the full version of WOGCL with WOGCL without OTL. Table \ref{tab: case study} demonstrates some dangling target entities are not correctly detected without OTL and become the nearest neighbor entities of source entities. It is obvious that the cosine similarity between a source entity and its nearest neighbor entity is slightly greater than the similarity between its counterpart. These dangling entities may be misaligned because of their high structural similarity to the source entities. In contrast, OTL can successfully detect those dangling entities with high probabilities. Besides, OTL makes matchable entity pairs more similar and prevents the appearance of other entities from being aligned. This case study demonstrates the effectiveness of WOGCL and the informativeness of OTL.

\section{Conclusion}
We proposed a novel weakly supervised framework WOGCL for entity alignment with dangling cases, which avoids the use of additional side information and enhances the generalizability of the entity alignment framework. By adopting both intra- and inter-attention mechanisms, we model the graph relations smartly. To replace the inefficient negative sampling strategy and improve the differentiation of entity representation, we propose \emph{Contrastive Learning with Hard Negative
Samples} to handle false negatives in a principled way and accelerate the convergence speed. We propose to utilize \emph{Opitmal Transport Learning} to lead GNN encoder to learn a more accurate mapping between source KG to target KG. To assist the inference of alignments on sparse datasets, we introduce \emph{Higher-order Similarity} to explore graph structure similarity via random walk. These four modifications enable the proposed WOGCL to achieve the SOTA performance on both \emph{consolidated evaluation} and \emph{relaxed evaluation}. The main experiments on DBP2.0 and MedED demonstrate that our method is able to beat all baselines. In addition, we verify the effectiveness of each component of our proposed methods by designing auxiliary experiments.

\begin{acks}
To Robert, for the bagels and explaining CMYK and color spaces.
\end{acks}

\bibliographystyle{ACM-Reference-Format}
\bibliography{sample-base}


\end{document}